\documentclass{article}

\usepackage[preprint,nonatbib]{neurips_2021}

\usepackage[utf8]{inputenc} 
\usepackage[T1]{fontenc}    
\usepackage{hyperref}       
\usepackage{url}            
\usepackage{booktabs}       
\usepackage{amsfonts}       
\usepackage{nicefrac}       
\usepackage{microtype}      
\usepackage{xcolor}         

\usepackage{amsmath,amssymb,amsfonts}
\usepackage{float}
\usepackage{multirow}
\usepackage{epsfig}
\usepackage{graphicx}
\usepackage{algorithm}
\usepackage{algorithmic}
\usepackage{subfigure}
\usepackage{epstopdf}
\usepackage{makecell}
\usepackage{bm}
\usepackage{wrapfig}
\usepackage[capitalize,noabbrev]{cleveref}
\pdfoutput=1
\title{Spiking GATs: Learning Graph Attentions via Spiking Neural Network}
\author{%
	Beibei Wang and Bo Jiang\thanks{Corresponding author (jiangbo@ahu.edu.cn).} \\
	School of Computer Science and Technology, Anhui University, Hefei, China
}

\begin{document}

\maketitle

\begin{abstract}
 Graph Attention Networks (GATs) have been intensively studied and widely used in graph data learning tasks. Existing GATs generally adopt the self-attention mechanism to conduct graph edge attention learning, requiring expensive computation. It is known that Spiking Neural Networks (SNNs) can perform inexpensive computation by transmitting the input signal data into discrete spike trains and can also return sparse outputs. Inspired by the merits of SNNs, in this work, we propose a novel Graph Spiking Attention Network (GSAT) for graph data representation and learning. In contrast to  self-attention mechanism in existing GATs, the proposed GSAT adopts a SNN module architecture which is obvious energy-efficient. Moreover, GSAT can return sparse attention coefficients in natural and thus can perform feature aggregation on the selective neighbors which makes GSAT perform robustly w.r.t graph edge noises. Experimental results on several datasets demonstrate the effectiveness, energy efficiency and robustness of the  proposed GSAT model.
\end{abstract}

\section{Introduction}
Graph Neural Networks (GNNs) have been shown powerfully on  dealing with  structured-graph data representation and learning.
In GNNs, each node updates its representation by recursively aggregating the messages of its neighbors~\cite{kipf2017semi,hamilton2017inductive,GeislerZG20,GeCN}. 
%
%
%
Inspired by the effectiveness of self-attention mechanism in computer vision and machine learning fields~\cite{chaudhari2021attentive, vaswani2017attention}, various self-attention
mechanisms have also been introduced in GNNs. 
Veli{\v c}kovi{\'c} et al.~\cite{velickovic2018graph} propose Graph Attention Networks (GAT) by absorbing neighbors' information via adaptive edge weights based on self-attention mechanism.
Wang et al.~\cite{wang2019heterogeneous} propose Heterogeneous Graph Neural Network (HAN) by jointly using node and semantic-level attention mechanisms. 
He et al.~\cite{cat} propose to learn conjoint attentions (CATs) by incorporating node features and structural interventions into attention learning.
Kim et al.~\cite{kim2021how} present a self-supervised graph attention network (SuperGAT) to estimate edges by utilizing two attention forms compatible with self-supervised tasks. 
Wang et al.~\cite{ijcai2021-425} propose Multi-hop Attention Graph Neural Network (MAGNA) by introducing multi-hop context information into attention learning.
Ye et al.~\cite{SGAT} present a sparse attention network (SGAT) to obtain sparse attention
coefficients by using $\ell_0$ norm regularization on attention coefficients. 
Shaked et al.~\cite{gatv2} propose a novel graph attention variant (GATv2) by adjusting the order of linear mapping and non-linear activation in  attention computation operations. 

After reviewing existing GATs, we can observe that existing models generally suffer from the following two limitations:
(A) Existing methods generally adopt the self-attention mechanism to compute
attention coefficients for all graph edges which generally require expensive computation. 
(B) Existing graph attention models return dense attention coefficients over all edges for message aggregation which thus are often sensitive to the noisy edges in the graph, such as incorrect or undesired redundant connections. 
To deal with the above limitation (B), sparse GATs~\cite{SGAT,neural_sparse,GeCN} have been proposed to learn a sparsely connected graph by introducing additional mask indication. However, the optimization of sparse GATs usually brings higher computational burden. 
Therefore, it is natural to raise a question: \textbf{beyond regular self-attention mechanism, can we learn the graph attentions efficiently and robustly by employing some other neural network mechanisms ? }

Inspired 
by recent works on Spiking Neural Networks (SNNs)~\cite{ghosh2009spiking,pmlr-v162-na22a,spikgnn}, in this paper, we propose to leverage SNN for learning graph attention and propose a novel \textbf{G}raph \textbf{S}piking \textbf{A}Ttention network (GSAT) for graph data representation and learning. 
The main advantages of the proposed GSAT are two aspects. 
First, GSAT can perform very \textbf{inexpensive} computation because SNN transmits the input data into discrete spike trains and thus conducts attention computation in the discrete trains space. 
Second, since SNN returns sparse outputs~\cite{ghosh2009spiking,tavanaei2019deep}, GSAT can naturally generate sparse attention coefficients and thus conduct message aggregation on the selective neighbors which performs robustly w.r.t graph attacks and noises. 
Overall, we summarize the main contributions as follows:
\begin{itemize}
	\item We propose a novel spiking attention mechanism based on SNN to learn graph edge attention sparsely and energy-efficiently. 
	\item Based on the proposed spiking attention mechanism, we develop a novel efficient Graph Spiking Attention Network (GSAT) for robust graph data representation  and learning.
	\item Experimental results on several datasets demonstrate the effectiveness, efficiency and robustness of the proposed GSAT model. 
\end{itemize}

\section{Related Works}
\subsection{Spiking Neural Networks}
Spiking Neural Networks (SNNs) have been widely studied in computer vision field mainly due to its energy-efficient network architecture~\cite{ghosh2009spiking,ponulak2011introduction,pfeiffer2018deep, tavanaei2019deep}.
For example, 
Cordone et al.~\cite{Cordone_2021_IJCNN} propose to adopt spiking neural networks to efficiently deal with the asynchronous and sparse data from event cameras.  
Zhu et al.~\cite{spikgnn} propose to combine SNN and GCN~\cite{kipf2017semi}  to learn graph node representation with obviously low cost.
It is known that, the key aspect of SNNs is to perform information propagation and transformation on the discrete spike trains with low energy cost. 
The spiking neuron mainly contains three operations: neuronal charging, neuronal firing and neuronal reset~\cite{vreeken2003spiking,IFreview,fang2021incorporating}, as briefly reviewed below. 
SNNs adopt an iterative signal propagation and transformation procedure. 
In each iteration, let $\mathbf{X}^{(t)}$ and $\mathbf{V}^{(t-1)}$ be the input and membrane potential at step $t$. Without loss of generality, the neuronal charging operation $\mathcal{C}(\cdot)$ can be defined as follows~\cite{vreeken2003spiking,fang2021incorporating},
\begin{align}\label{Eq:charge}
\mathbf{V}^{(t)} = \mathcal{C}(\mathbf{V}^{(t-1)}, \mathbf{X}^{(t)})
\end{align}
where $t = 1, 2,\cdots, T$ denotes the iterative time step. $\mathbf{V}^{(t)}$ is the membrane potential after neuronal charging and $\mathbf{V}^{(0)}=\bm{0}$. $\mathcal{C}(\cdot)$ denotes a specific charging function and different spiking neuron models can have different forms of charging functions.  

After neuronal charging step, SNNs then adopt a neuronal firing operation $\mathcal{F}(\cdot)$ at iteration $t$ which can be  described in general as follows~\cite{vreeken2003spiking,fang2021incorporating}
\begin{align}\label{Eq:fire}
\mathbf{S}^{(t)} = \mathcal{F}(\mathbf{V}^{(t)}-\mu)
\end{align}
where $\mu$ is the firing threshold and $\mathbf{S}^{(t)}$ represents the output spikes after neuronal firing. $\mathcal{F}(\cdot)$ is usually defined as the heaviside step function~\cite{fang2021deep} as,
\begin{equation}
\mathcal{F}(x)=\left\{
\begin{array}{rcl}
1 \,\,\,\,\,\,\,\,      &      & {x \geq 0}\,,\\
0 \,\,\,\,\,\,\,\,    &      & \mathrm{otherwise\,.}
\end{array} \right.
\end{equation}
After firing spikes, the spiking neuron needs to reset the membrane potential and the reset function $\mathcal{R}(\cdot)$ can be formulated as follows~\cite{han2020rmp},
\begin{align}\label{Eq:reset}
\mathbf{V}^{(t)} = \mathcal{R}(\mathbf{V}^{(t)}, \mu) 
\end{align}
where $\mathbf{V}^{(t)}$ is the membrane potential after firing spikes and used for the next iterative step. 
In many SNNs, one can simply define $\mathcal{R}(\mathbf{V}^{(t)}, \mu)$ as 
$\mathcal{R}(\mathbf{V}^{(t)}, \mu) =\mathbf{V}^{(t)} - \mu$, as suggested in works~\cite{ledinauskas2020training,feng2022building}. 

\subsection{Graph Attention Networks}
Graph Attention Networks (GATs)~\cite{attention_survey,velickovic2018graph,gatv2} aim to employ self-attention mechanism for message passing in graph neural network. 
For example, in work GAT~\cite{velickovic2018graph}, it uses a self-attention mechanism to assign different weights for edges and then conducts information aggregation on the learned  weighted graph. 
To be specific, given a graph  $G(\mathbf{A}, \mathbf{X})$ where
$\mathbf{A}=[a_{ij}]\in \mathbb{R}^{n\times n}$ denotes the adjacency matrix and $\mathbf{X}=[\boldsymbol{x}_1, \boldsymbol{x}_2\cdots \boldsymbol{x}_n] \in \mathbb{R}^{n\times d}$ represents the feature collection of graph nodes. 
GAT~\cite{velickovic2018graph} first adopts self-attention module with the weight vector $\theta\in \mathbb{R}^{1 \times 2d'}$ to compute the attention coefficients for edges as follows,
\begin{align}\label{Eq:att}
& \boldsymbol{h}_{j} = \boldsymbol{x}_j \mathbf{W}, \,\,\,\forall j \in \mathcal{N}_i\cup i
\\
&  \bm{\alpha}_{ij}=[\boldsymbol{h}_{i}\|\boldsymbol{h}_{j}] {\theta}^{\rm{T}}
\end{align}
where $\|$ denotes the concatenation operation. $\mathbf{H}=[\boldsymbol{h}_1, \boldsymbol{h}_2\cdots \boldsymbol{h}_n] \in \mathbb{R}^{n\times d'}$ are transformed graph node representations and $\mathbf{W} \in \mathbb{R}^{d \times d'}$ denotes the learnable parameter of linear projection.   $\mathcal{N}_i$ denotes the neighborhood set of node $v_i$.
GAT also  normalizes  the attention coefficients by using Softmax operation, %
\begin{equation}\label{Eq:att1}
\bm{\alpha}'_{ij} = \frac{\exp\big(\text{LeakyReLU}(\bm{\alpha}_{ij})\big)}{\sum\nolimits_{j\in \mathcal{N}_i}\exp\big(\text{LeakyReLU}(\bm{\alpha}_{ij})\big)}
\end{equation}
where $\text{LeakyReLU}$ is the non-linear activation function.
Based on the above attention coefficients, GAT~\cite{velickovic2018graph} then conducts neighborhood aggregation as follows,
\begin{equation}\label{Eq:gat}
\boldsymbol{h}^{'}_i = \sigma\big(\sum\nolimits_{j\in\mathcal{N}_i}\bm{\alpha}'_{ij}\boldsymbol{h}_j\big)
\end{equation}
where $\sigma(\cdot)$ denotes the activation function, such as ReLU and ELU. 
\begin{figure*}[!ht]
	\centering
	\centering
	\includegraphics[width=0.8\textwidth]{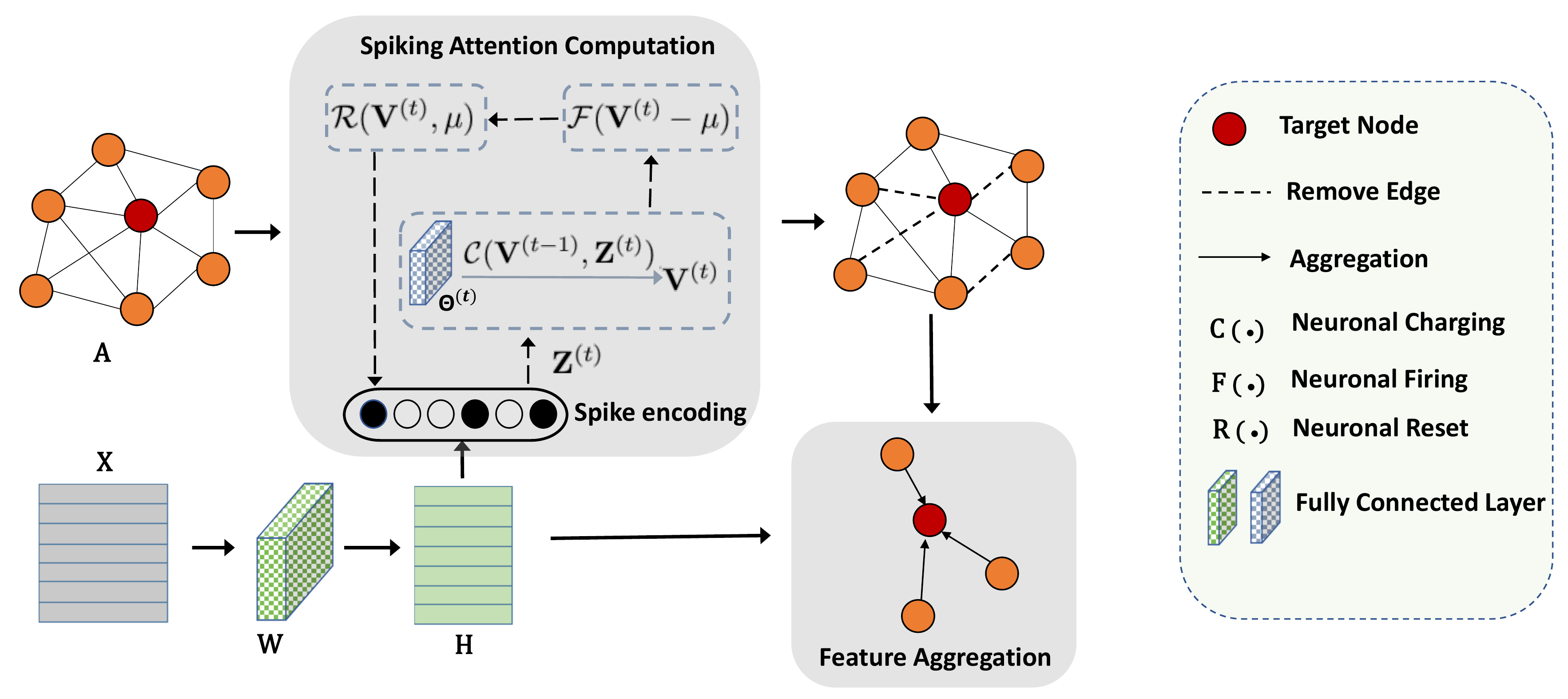}
	\caption{ The overall framework of our proposed Graph Spiking Attention Network. }\label{fig::model}
\end{figure*}

\section{Graph Spiking Attention Network}


In this section, we present our Graph Spiking Attention Network (GSAT) by leveraging SNN model for graph attention computation. 
In contrast to self-attention mechanism in GATs, SNNs use discrete spikes to propagate and transform the edge information, which thus require low energy cost and can also return the sparse outputs~\cite{tavanaei2019deep}.
Therefore, we propose to leverage attention learning based on SNN model to learn sparse attention coefficients.
Overall, the proposed GSAT mainly consists of two modules: Spiking Attention Computation and Feature Aggregation, as shown in Figure \ref{fig::model}.  

\subsection{Spiking Attention Computation}

The attention mechanism in Eq.(6) can be implemented by a single-layer feed-forward network with learnable parameter vector ${\theta}\in \mathbb{R}^{1\times2d'}$~\cite{velickovic2018graph}. 
We first split ${\theta}$ into two parts, i.e., $\theta_1 = \theta[1:d'] \in \mathbb{R}^{1\times d'}$ and $\theta_2 = \theta[d'+1:2d']\in \mathbb{R}^{1\times d'}$. 
Then, Eq.(6) can  be  rewritten as follows,
\begin{align}\label{Eq:att2}
\bm{\alpha}_{ij} ={\boldsymbol{h}_i} {\theta^{\rm{T}}_1} + {\boldsymbol{h}_j}{\theta^{\rm{T}}_2}
\end{align}
Let $\Theta = [\theta^{\rm{T}}_1 \| \theta^{\rm{T}}_2] \in \mathbb{R}^{d' \times 2}$,
and the compact matrix computation process of Eq.(\ref{Eq:att2}) can be implemented as follows,
\begin{align}\label{Eq:att3}
&\mathbf{S} =\mathbf{H}\Theta, \,\,\,\,  \mathbf{S} \in \mathbb{R}^{n \times 2}
\\
&\bm{\alpha} = \mathbf{S}[:,1]\textbf{1} + ({\mathbf{S}[:,2]}{\textbf{1}})^{\rm{T}}
\end{align}
where $\textbf{1}=(1,1,\cdots,1) \in \mathbb{R}^{1 \times n}$. 

Based on the above reformulation, we can adopt spiking neural networks to implement the graph attention learning which consists of Spiking encoding, Neuronal charging, Neuronal firing and neuronal reset~\cite{vreeken2003spiking,IFreview,fang2021incorporating,spikgnn}, as introduced below in detail. 
   
\noindent\textbf{Spiking encoding. }
From Eqs.(10,11), the attention scores are computed based on the transformed node representation $\mathbf{H}$. SNNs accept discrete inputs while $\mathbf{H}$ encodes continuous values. We thus need to convert the continuous node representations to the discrete spike trains. Specifically, let $\mathbf{P}^{(t)} = [p^{(t)}_{ij}]\in \mathbb{R}^{n \times d'}$ be the randomly generated  matrix where ${p}^{(t)}_{ij}\in[0,1]$ and $t$ denotes the iterative step. 
We then adopt a probability-based Poisson encoder with probability ${p}^{(t)}_{ij}$ to transmit the input data $\textbf{H}$ to the discrete spike trains~\cite{sharmin2020inherent,9556508}.
The transmitting operation is defined as follows,
\begin{equation}
\mathbf{Z}^{(t)}_{ij}=\left\{
\begin{array}{rcl}
1 \,\,\,\,\,\,\,\,      &      & {\mathbf{H}_{ij} \geq {p}^{(t)}_{ij}}\, ,\\
0 \,\,\,\,\,\,\,\,    &      & \mathrm{otherwise\,.}
\end{array} \right.
\end{equation}
where $\mathbf{Z}^{(t)}$ denotes the discrete spike trains at step $t$.

\noindent\textbf{Neuronal charging. }
After the above discrete spiking encoding, we can obtain the spike trains $\mathbf{Z}^{(t)}$ as the input of spiking neuron. 
Then, we adopt the spiking neuronal charging operation to implement attention computation. 
Specifically, we take the Integrate-and-Fire (IF) mechanism~\cite{IF, IFreview} as the spiking neuron and define the charging function $\mathcal{C}(\cdot)$ to compute attention scores as,  
\begin{align}\label{Eq:gat_charge}
\mathbf{V}^{(t)} = \mathcal{C}(\mathbf{V}^{(t-1)},\mathbf{Z}^{(t)} )=\mathbf{V}^{(t-1)}+\mathbf{Z}^{(t)}\Theta^{(t)}
\end{align}
where $\Theta^{(t)}\in \mathbb{R}^{d' \times 2}$ is a trainable parameter matrix. $\mathbf{V}^{(t-1)}$ is the membrane potential at time step $t$ and we simply set $\mathbf{V}^{(0)}=\bm{0}$.

\noindent\textbf{Neuronal firing. }
As mentioned above, SNNs return a discrete spike trains. Therefore,  
we use the spiking neuronal firing function $\mathcal{F}(\cdot)$ to conduct the discrete operation as
\begin{equation}\label{Eq:gat_fire}
\mathbf{S}^{(t)}=\mathcal{F}(\mathbf{V}^{(t)}-\mu)=\left\{
\begin{array}{rcl}
1 \,\,\,\,     &      & {\mathbf{V}^{(t)}-\mu \geq 0}\,\\
0 \,\,\,\,    &      & \mathrm{otherwise\,}
\end{array} \right.
\end{equation} 
where $\mathbf{S}^{(t)}$ represents the sparse spike trains returned by SNNs and $\mu$ is the threshold voltage of neuronal firing. 

\noindent\textbf{Neuronal reset. }
After the firing operation, we need to reset the membrane potential of spiking neuron. In this paper, we adopt the soft reset~\cite{SpikingJelly} to define the neuronal reset function $\mathcal{R}(\cdot)$ as 
\begin{align}\label{Eq:gat_reset}
\mathbf{V}^{(t)} = \mathcal{R}(\mathbf{V}^{(t)}, \mu) = \mathbf{V}^{(t)} - \mu
\end{align}
where $\mathbf{V}^{(t)}$ is the membrane potential for the next step.

\noindent\textbf{Attention estimation. }
By conducting operations Eqs.(12-15) $T$ times, we can obtain a series of sparse spike trains $\{\mathbf{S}^{(1)},\mathbf{S}^{(2)}\cdots\mathbf{S}^{(T)}\}$. We then utilize the average operation to output the final spike trains as, 
%
\begin{align}\label{Eq:gat_average}
\mathbf{S} = \frac{1}{T}\sum_{t=1}^{T}\mathbf{S}^{(t)}
\end{align}
Finally, we compute the  attentions according to Eq.(11) as
\begin{align}\label{Eq:spike_att}
\bm{\alpha} = \mathbf{S}[:,1]\textbf{1} + ({\mathbf{S}[:,2]}\textbf{1})^{\rm{T}}
\end{align}

To preserve the sparse property of the attention coefficients $\bm{\alpha}$, instead of softmax function, we use the symmetric normalization to normalize $\bm{\alpha}$  as
\begin{align}\label{Eq:sys_norm}
\bm{\alpha}'_{ij} = \frac{\bm{\alpha}_{ij}}  { \sqrt{\sum\limits_{j\in\mathcal{N}_i}{\bm{\alpha}_{ij}}}\sqrt{\sum\limits_{i\in\mathcal{N}_j}{\bm{\alpha}_{ij}}} }
\end{align}
The whole attention computation by our spiking attention mechanism is summarized in Algorithm 1.  
\begin{algorithm}[h]
	\caption{Spiking Attention Computation}
	\begin{algorithmic}[1]
		\STATE {\bfseries Input:} Node representation $\mathbf{H} \in \mathbb{R}^{n \times d'}$, parameter $\mu$ 
		\STATE Initialize $\mathbf{V}^{(0)}=\bm{0}$
		\FOR{$t=1$ {\bfseries to} $T$}
		\STATE Spiking encoding : 
		$\mathbf{Z}^{(t)}\leftarrow \text{PoissonEncoder}(\mathbf{H}) 
		$
		\STATE Neuronal charging :  
		$\mathbf{V}^{(t)} \leftarrow \mathbf{V}^{(t-1)} + \mathbf{Z}^{(t)}\Theta^{(t)}$
		\STATE Neuronal firing :  
		$\mathbf{S}^{(t)} \leftarrow \mathcal{F}(\mathbf{V}^{(t)}-\mu)$
		\STATE Neuronal reset :  
		$\mathbf{V}^{(t)} \leftarrow \mathbf{V}^{(t)}-\mu$
		\ENDFOR 
		\STATE Compute attention  matrix by Eqs.(\ref{Eq:gat_average},\ref{Eq:spike_att})
		\STATE Normalize attention matrix by Eq.(\ref{Eq:sys_norm})
		\emph{}
	\end{algorithmic}
\end{algorithm}

\subsection{Feature Aggregation}

Based on the proposed spiking attention computation, we can obtain sparse graph attention coefficients $\bm{\alpha}_{ij}$ and then conduct the feature aggregation as follows,
\begin{equation}\label{Eq:fa}
\boldsymbol{h}'_i =  \sigma\big(\sum_{j\in\mathcal{N}_i}\bm{\alpha}'_{ij}\boldsymbol{h}_j\big)
\end{equation}
Also, by using multi-head mechanism~\cite{velickovic2018graph},  the feature aggregation  
of the proposed GSAT can finally be formulated as, 
\begin{equation}\label{Eq:layer_fa}
\boldsymbol{h}^{'}_i = {\Big\|}^{K}_{k=1} \sigma\big(\sum_{j\in\mathcal{N}_i}{\bm{\alpha}'}^{(k)}_{ij}\boldsymbol{h}^{(k)}_j\big)
\end{equation}
where $k=1,2,\cdots,K$ denotes the $k$-th head. 

Similar to traditional GAT architecture~\cite{velickovic2018graph}, we adopt the multi-layer neural network architecture to learn graph node representation for the down-stream tasks, such as semi-supervised learning, clustering, etc. 
For node classification task, 
one can employ the softmax operation on the last layer to obtain the final predications $\mathbf{O}\in \mathbb{R}^{n \times c}$ and then compute the cross-entropy loss as~\cite{kipf2017semi}, 
\begin{equation}
\mathcal{L}_{\textrm{Semi}} = -\sum\limits_{i\in \mathbb{L}} \sum^c\limits_{j=1} \textbf{Y}_{ij}\mathrm{ln} \textbf{O}_{ij}
\end{equation}
where $i\in \mathbb{L}$ indicates the $i$-th labeled node and $\textbf{Y}_{i\cdot}$ of $\textbf{Y}$ denotes the ground-truth label of node $v_i$. 
We use Glorot initialization~\cite{glorot2010understanding} to initialize network and use the commonly used Adam algorithm~\cite{Adam} to train network.

\section{Experiment}
\subsection{Evaluation Datasets}
To validate the effectiveness of the proposed GSAT, we use five benchmark datasets including Cora, Citeseer, Pubmed~\cite{sen2008collective}, Amazon Photo and Amazon Computers~\cite{shchur2018pitfalls}. 
Cora, Citeseer and Pubmed are citation datasets in which nodes indicate the documents and edges encode the link relationships between documents. 
Amazon Photo and Amazon Computers are co-purchase datasets in which nodes indicate the products and edge exists between two products if they are often purchased together.  
Table \ref{dataset} summarizes the usage of all datasets used in our experiments.
\begin{table*}[!htpb]
	\centering
	\caption{The brief usage of all datasets.}
	\centering
	\begin{tabular}{c c c c c c }
		\hline
		\hline
		Dataset & Cora & Citeseer & Pubmed  & Amazon Photo & Amazon Computers \\
		\hline
		Nodes  & 2708 & 3327 & 19717 & 7487 &13381\\
		Feature   & 1433 & 3703 & 500 & 745 &767 \\		
		Edges    & 5429 & 4732 & 44338 & 119043 &245778\\
		Classes & 7 & 6 & 3 & 8 & 10 \\
		\hline
		\hline
	\end{tabular}
	\label{dataset}
\end{table*}

\subsection{Experimental Settings}
\noindent\textbf{Data setting. }
For citation datasets, we randomly sample $20$ labeled nodes per class as training data, $500$ and $1000$ nodes as validation and test data respectively.
For co-purchase datasets, we follow the work~\cite{shchur2018pitfalls} and randomly sample $20$ labeled nodes per class as training data, $30$ nodes per class as validation data, and the rest nodes as testing data.

\noindent\textbf{Parameter setting. }
Similar to the previous GATs~\cite{velickovic2018graph}, our proposed GSAT involves two-layer network architecture with multi-head spiking attention mechanism.
We set the number of hidden-layer units and heads to $\{8,8\}$.
The weight decay is set to $5e-4$ for citation datasets and  $5e-5$ for co-purchase dataset, respectively.
The parameter $\mu$ in the SNN module is determined based on the validation set.
We train our model by using widely-used Adam algorithm~\cite{Adam} and initialize the network parameters by
using Glorot initialization~\cite{glorot2010understanding}.

\subsection{Comparison Results}
We take Graph Convolutional Network (GCN)~\cite{kipf2017semi}, Graph Attention Network (GAT)~\cite{velickovic2018graph}, GAT-Topk~\cite{velickovic2018graph}, DropEdge~\cite{dropedge}, Sparse Graph Attention Network (SGAT)~\cite{SGAT} and Graph elastic Convolutional Network (GeCN)~\cite{GeCN} as compared methods.
For GAT-Topk, we adopt GAT~\cite{velickovic2018graph} to obtain sparse attention matrix by only using the top-k attention scores.
All results are obtained by running the original codes provided by authors. 
Table \ref{ncla1} reports the node classification performance of the five datasets. 
Overall, our proposed GSAT achieves the best performance on all datasets. Moreover, we can have the following observations: 
(1) GSAT performs best on all datasets which obviously shows the effectiveness of GSAT on boosting GATs' learning performance.
(2) GSAT outperforms DropEdge~\cite{dropedge} which uses random sparse graph structure. This indicates that our GSAT can conduct neighbor selection more effectively for layer-wise aggregation.
(3) GSAT obtains higher performance than GAT-Topk~\cite{velickovic2018graph}, SGAT~\cite{SGAT} and GeCN~\cite{GeCN} which also adopt sparse attention/graph structure learning strategies. This further demonstrates the merit of the proposed SNN based sparse attention learning module. 
\begin{table*}[!ht]
	\normalsize
	\centering
	\caption{\upshape Node classification performance of five datasets.
	}
	\begin{tabular}{c|| c |c |c |c| c }
		\hline
		\hline		
		Method & Cora & Citeseer & Pubmed &Amazon Photo &Amazon Computers\\
		\hline
		GCN  & 81.27$\pm$1.08  &69.92$\pm$1.23 & 80.21$\pm$1.26 & 89.92$\pm$1.02 & 81.27$\pm$2.20\\
		DropEdge  & 82.04$\pm$1.34  & 70.22$\pm$1.15 & 80.74$\pm$1.10 &90.34$\pm$0.74  &82.10$\pm$1.74 \\
		GAT  &82.86$\pm$0.94  &71.23$\pm$1.09
		&80.26$\pm$1.10 & 90.20$\pm$0.85 & 81.50$\pm$1.55 \\
		GAT-Topk  & 82.40$\pm$1.42  &71.03$\pm$1.21  &80.93$\pm$1.36  &90.40$\pm$0.96  &81.96$\pm$1.33 \\
		GeCN  & 83.24$\pm$1.31  &71.34$\pm$1.44  &81.18$\pm$1.24  & 90.81$\pm$1.37 & 82.31$\pm$2.14 \\
		SGAT  &81.90$\pm$0.96  &71.22$\pm$0.72 &81.20$\pm$1.45 & 91.15$\pm$0.60 & 82.42$\pm$1.71 \\
		\hline
		GSAT  &\textbf{84.18$\pm$1.00}  &\textbf{72.90$\pm$0.78} &\textbf{82.39$\pm$0.78} & \textbf{92.22$\pm$0.58} & \textbf{83.91$\pm$1.34} \\
		\hline
		\hline
	\end{tabular}
	\label{ncla1}
\end{table*}

\subsection{Robustness Evaluation}
\begin{figure*}[!ht]
	\centering
	\subfigure[Nettack attack]
	{
		\begin{minipage}[t]{0.4\textwidth}
			\centering
			\includegraphics[width=5.5cm]{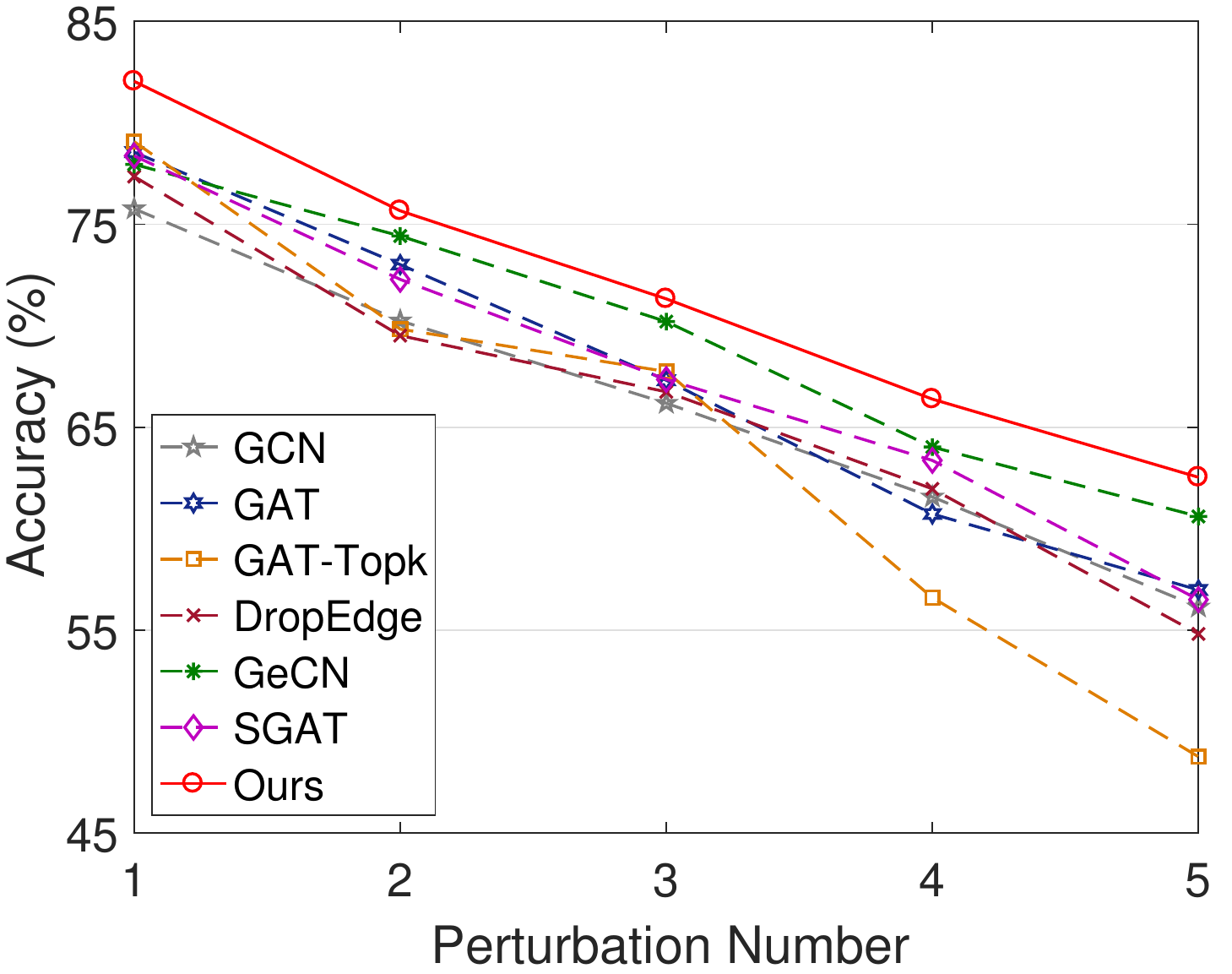}
		\end{minipage}
	}
	\subfigure[Random attack]
	{
		\begin{minipage}[t]{0.4\textwidth}
			\centering
			\includegraphics[width=5.5cm]{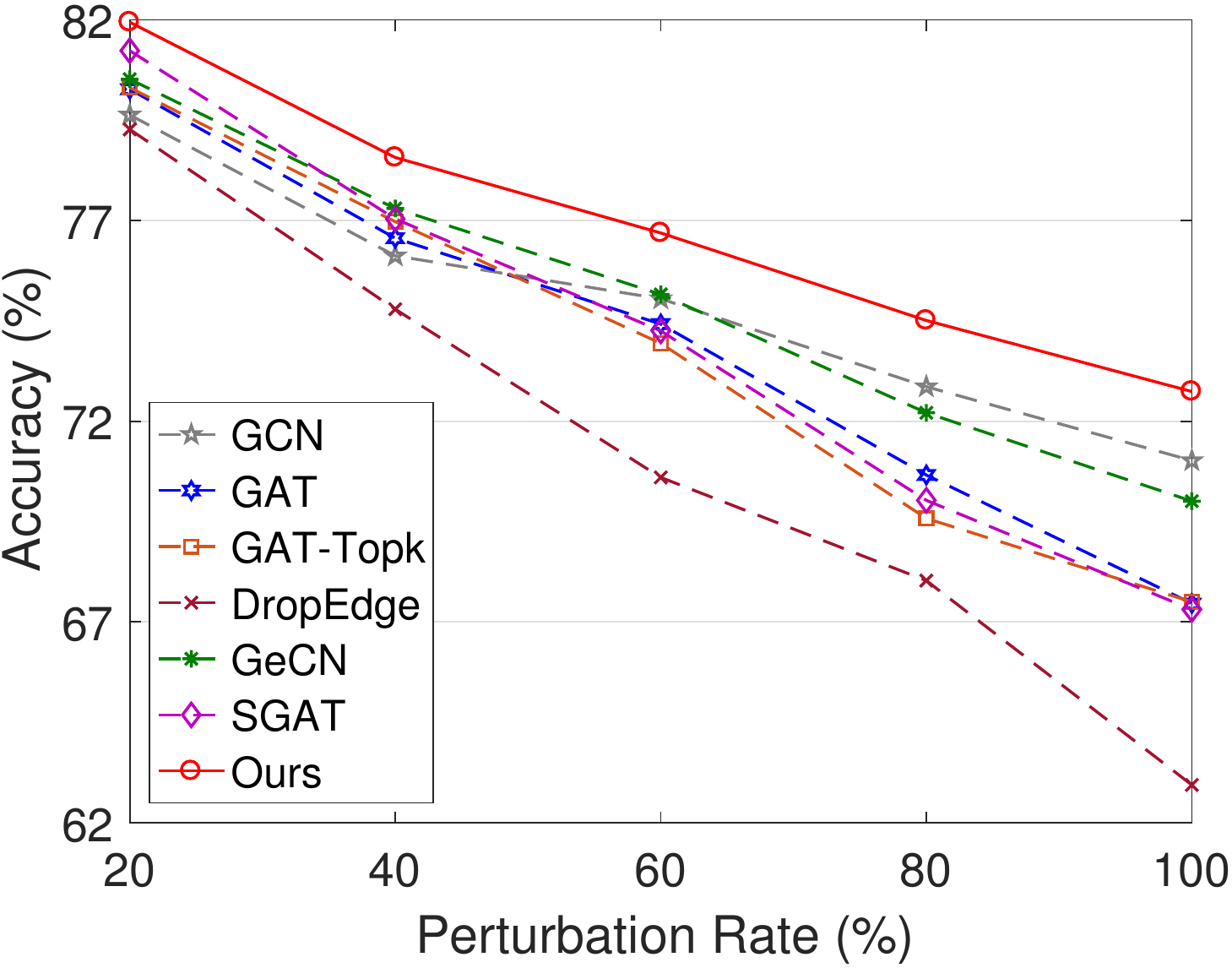}
		\end{minipage}
	}
	\caption{Results of Comparison methods on Cora dataset under Nettack and Random attack. }\label{fig::cora}
\end{figure*}

\begin{figure*}[!ht]
	\centering
	\subfigure[Nettack attack]
	{
		\begin{minipage}[t]{0.4\textwidth}
			\centering
			\includegraphics[width=5.5cm]{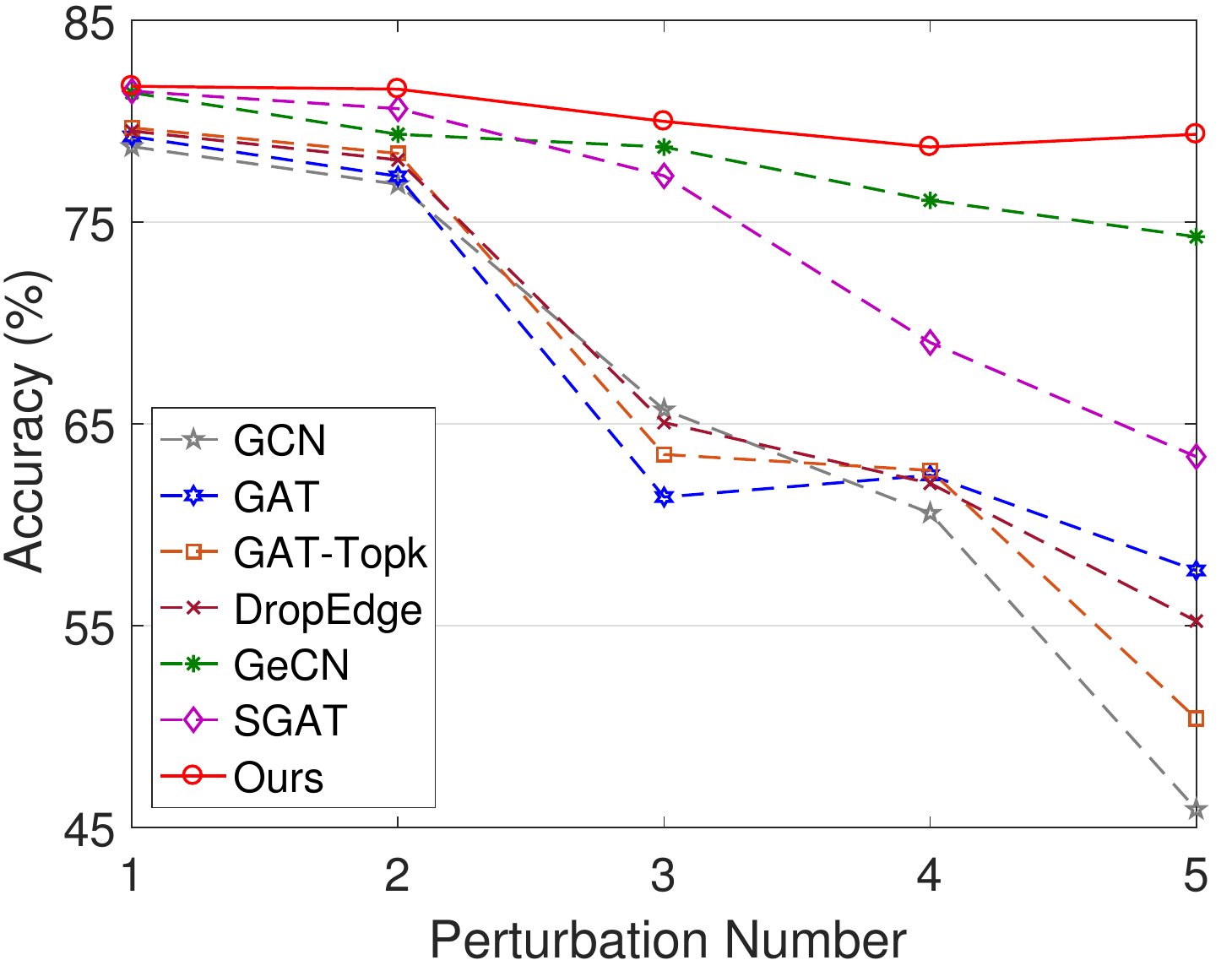}
		\end{minipage}
	}
	\subfigure[Random attack]
	{
		\begin{minipage}[t]{0.4\textwidth}
			\centering
			\includegraphics[width=5.5cm]{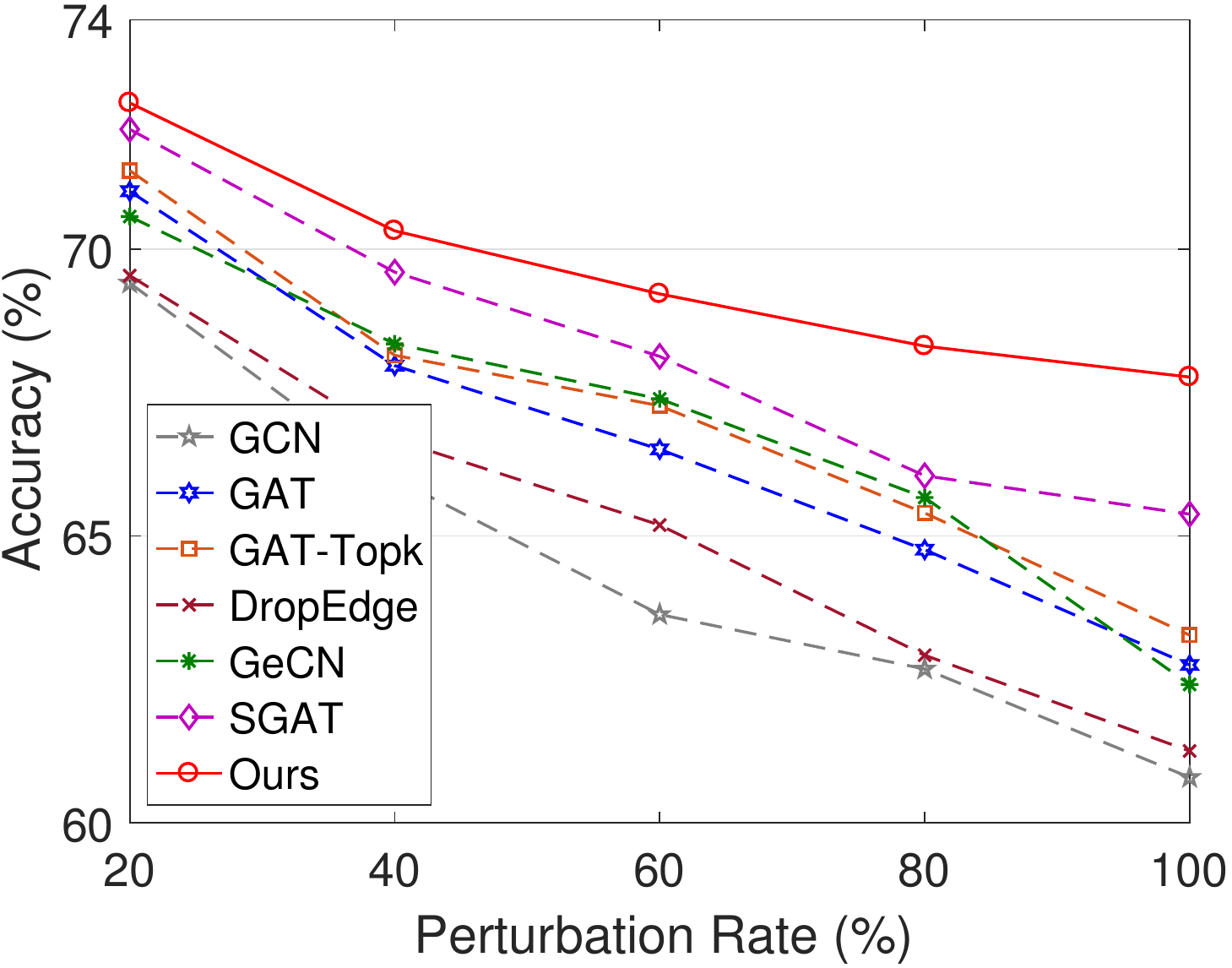}
		\end{minipage}
	}
	\caption{Results of Comparison methods on Citeseer dataset under Nettack and Random attack. }\label{fig::citeseer}
\end{figure*}

To evaluate the robustness of GSAT model, we use the structural perturbation Nettack ~\cite{zugner2018adversarial} and Random attack~\cite{li2020deeprobust} on Cora and Citeseer datasets. 
For Nettack method, we generate the noisy graph for targeted nodes with the perturbation number varying from $1$-$5$.
For Random method, we generate the noisy graph by adding some edges with the perturbation rate varying from $20\%$-$100\%$. 
We follow the data splits provided by work~\cite{prognn} and report the comparison results on Cora dataset in Figure \ref{fig::cora} and Citeseer dataset in Figure \ref{fig::citeseer}.
From Figs \ref{fig::cora}-\ref{fig::citeseer}, one can observe that (1) GSAT obtains obviously higher classification performance which shows the effectiveness of GSAT on enhancing model's robustness. 
(2) GSAT performs more robustly than SGAT~\cite{SGAT} and GeCN~\cite{GeCN} which indicates the benefit of sparse attention learning on enhancing the robustness of the proposed GSAT. 
%
%
%
\subsection{Efficiency Evaluation} 
\begin{figure}[!htp]
	\centering
	\centering
	\includegraphics[width=0.4\textwidth]{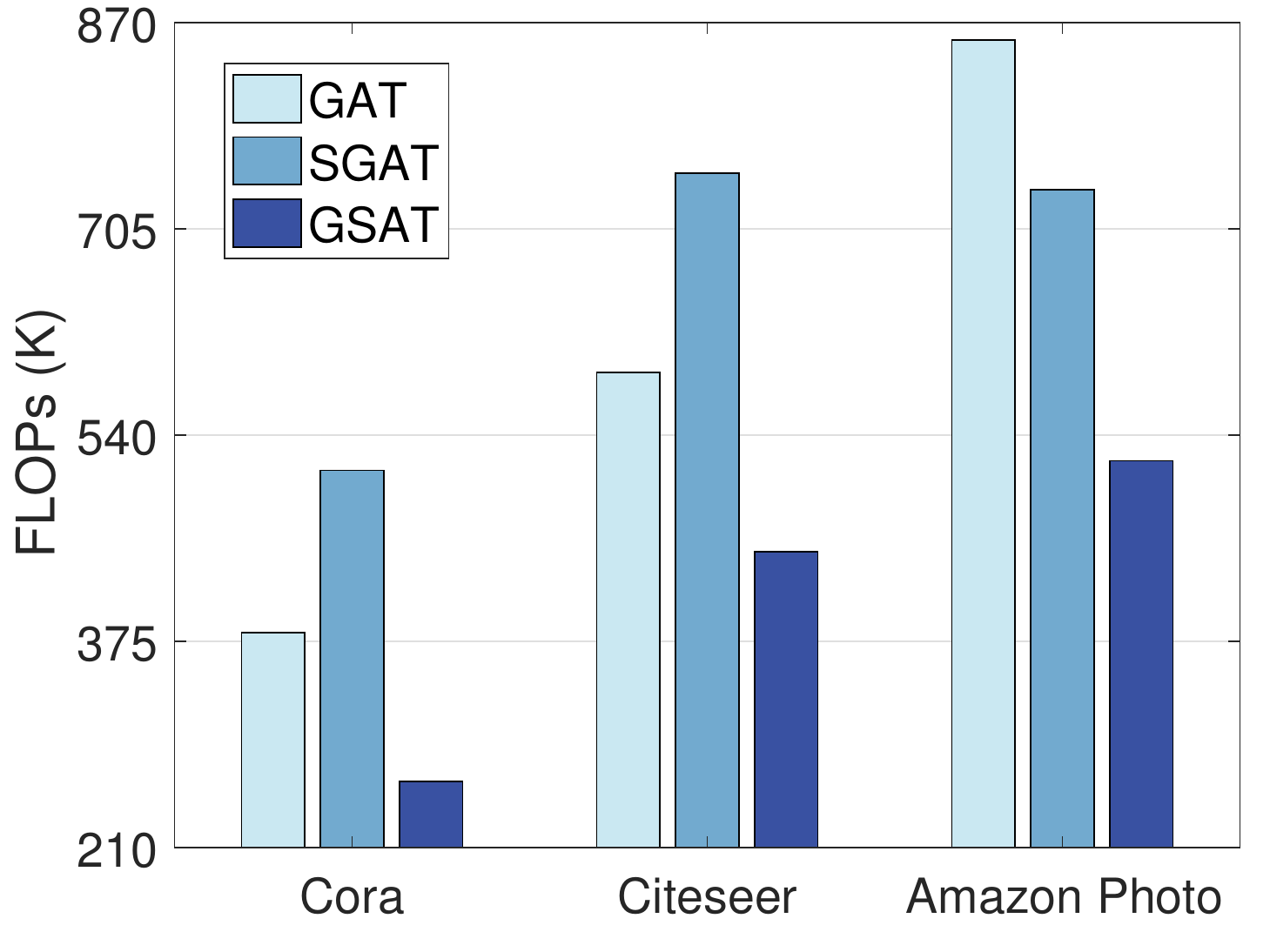}
	\caption{Comparisons of FLOPs on Cora, Citeseer and Amazon Photo datasets. }\label{fig::flops}
\end{figure}

To evaluate the energy efficiency of our GSAT, we take float operations (FLOPs)~\cite{hunger2005floating} as the measurement and compare our GSAT with some related GATs including original GAT and sparse GAT (SGAT).  Figure \ref{fig::flops} shows the comparison results. We can note that GSAT can perform attention learning and feature aggregation with obviously less energy cost, indicating the energy efficiency of the proposed GSAT on conducting graph data learning tasks. 
%
\begin{figure*}[!ht]
	\centering
	\subfigure[Parameter analysis on Cora dataset]
	{
		\begin{minipage}[t]{0.4\textwidth}
			\centering
			\includegraphics[width=5cm]{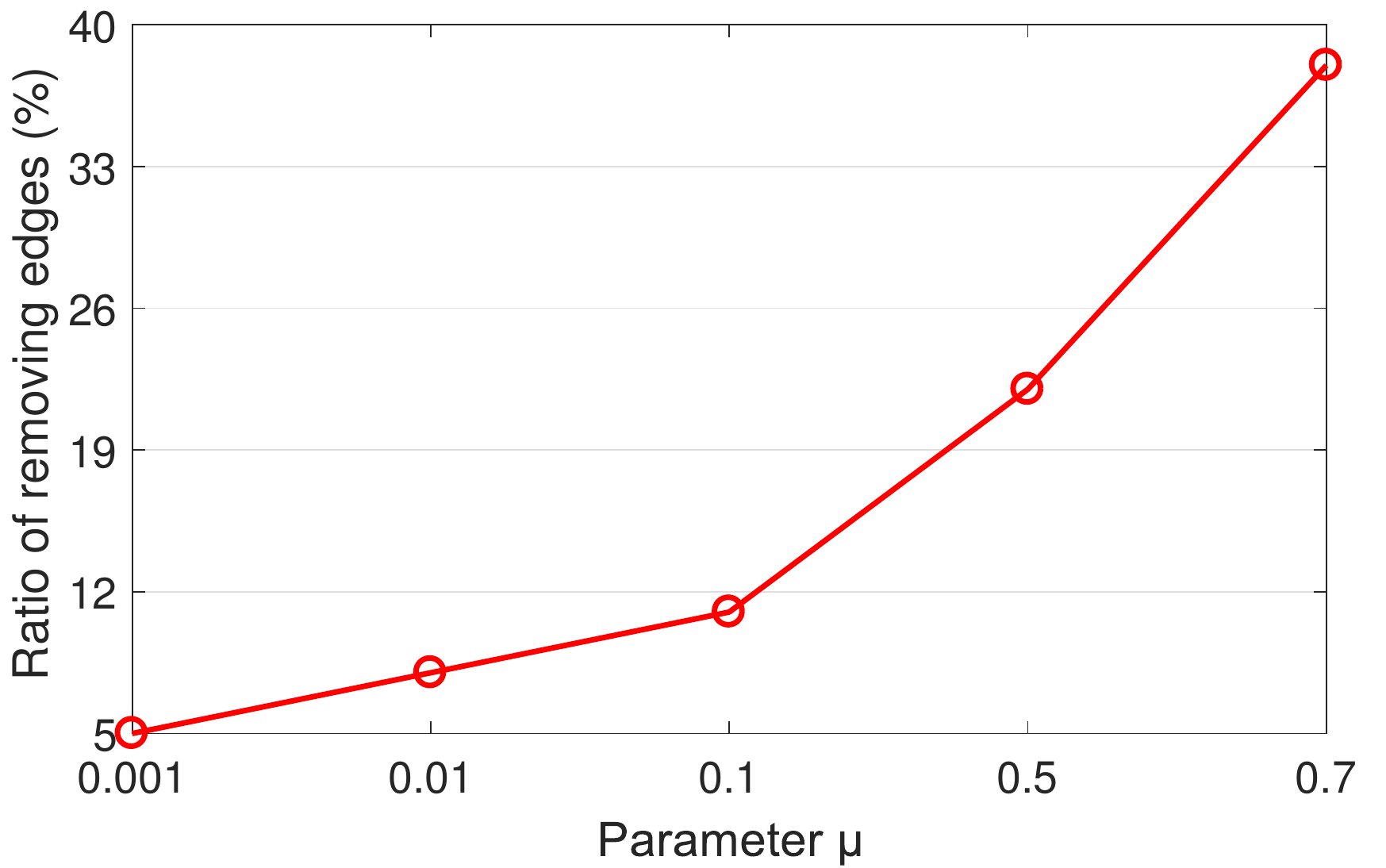}
		\end{minipage}
		\begin{minipage}[t]{0.4\textwidth}
			\centering
			\includegraphics[width=5cm]{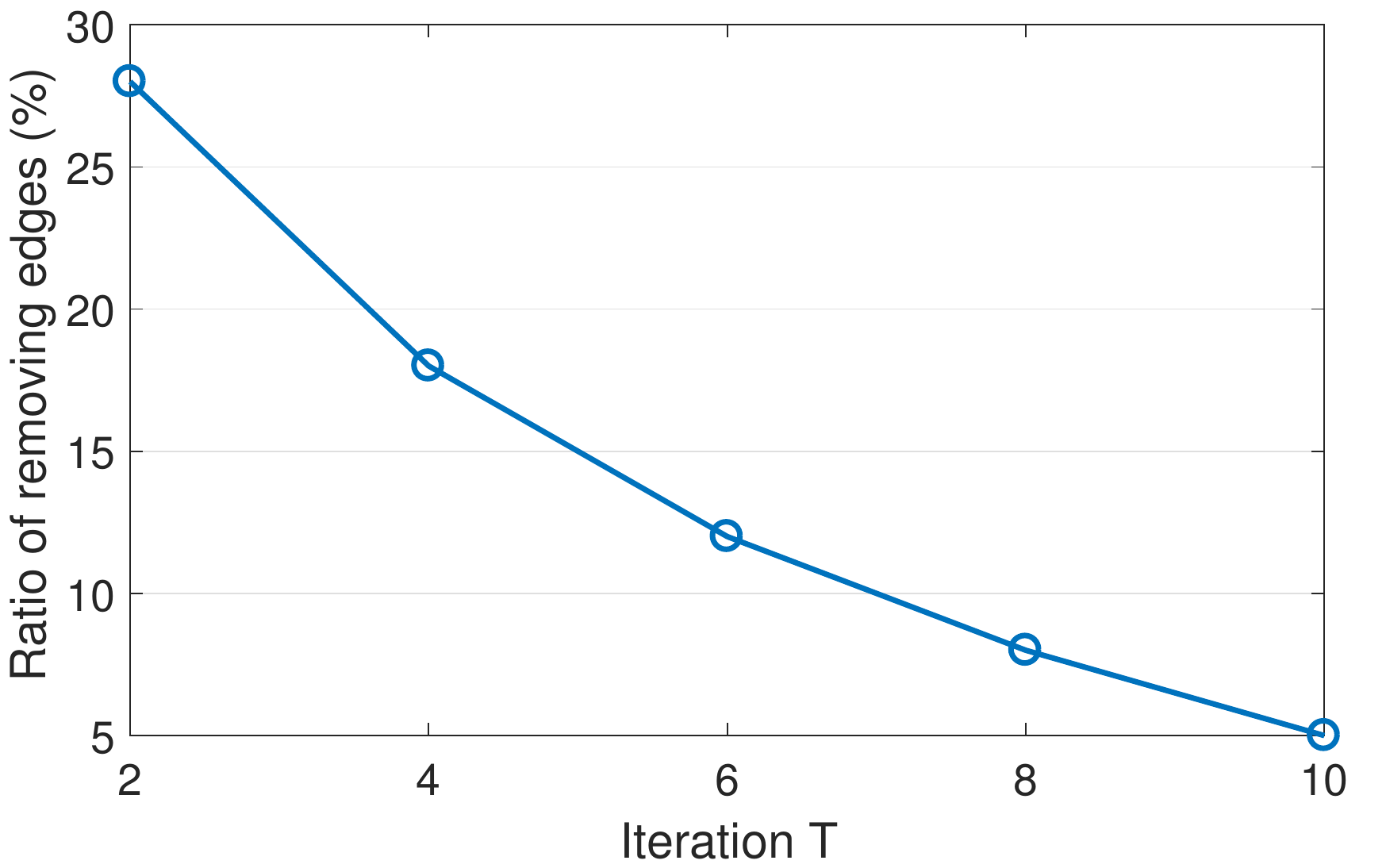}
		\end{minipage}	
	}
	\subfigure[Parameter analysis on Citeseer dataset]
	{
		\begin{minipage}[t]{0.4\textwidth}
			\centering
			\includegraphics[width=5cm]{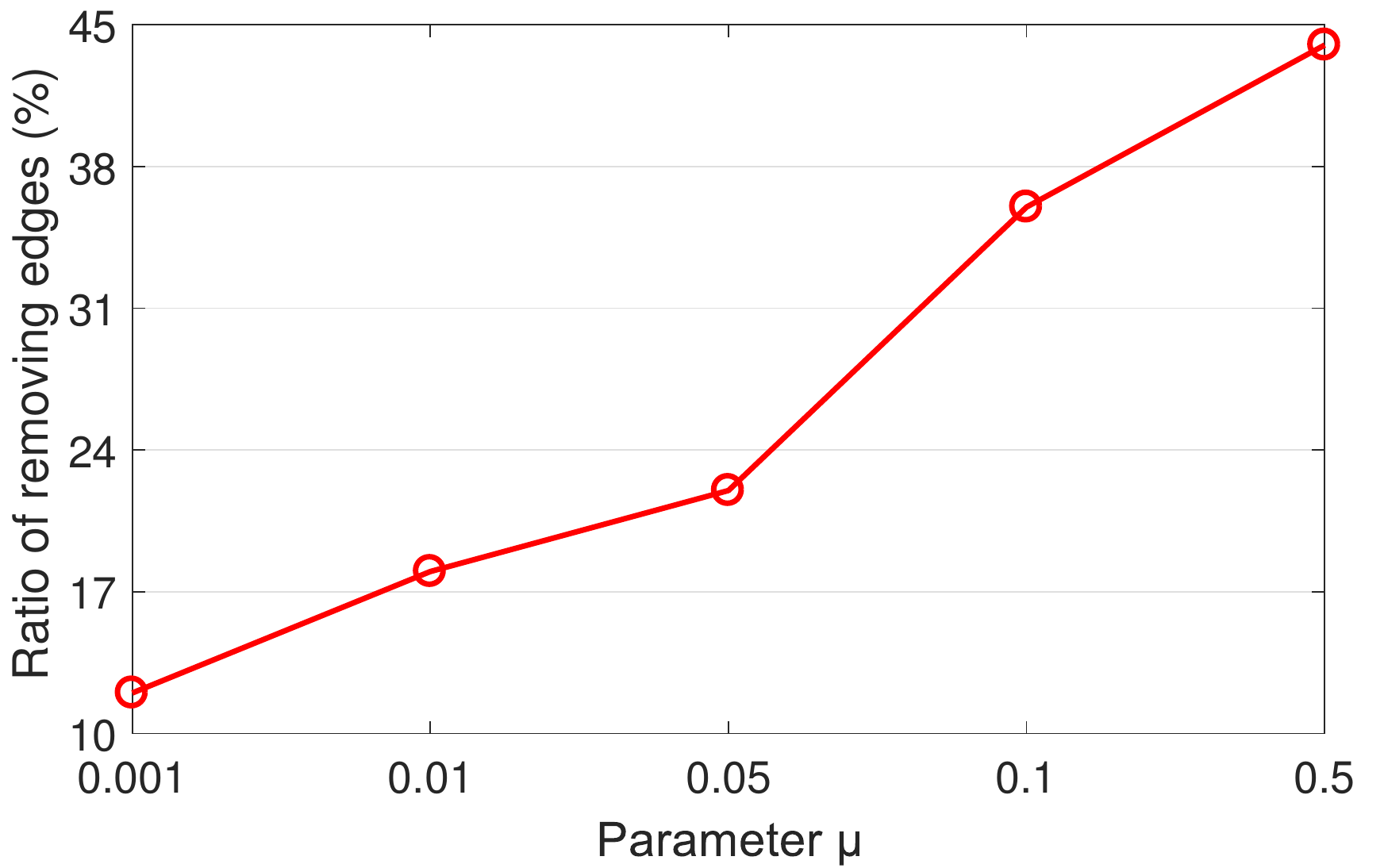}
		\end{minipage}
		\begin{minipage}[t]{0.4\textwidth}
			\centering
			\includegraphics[width=5cm]{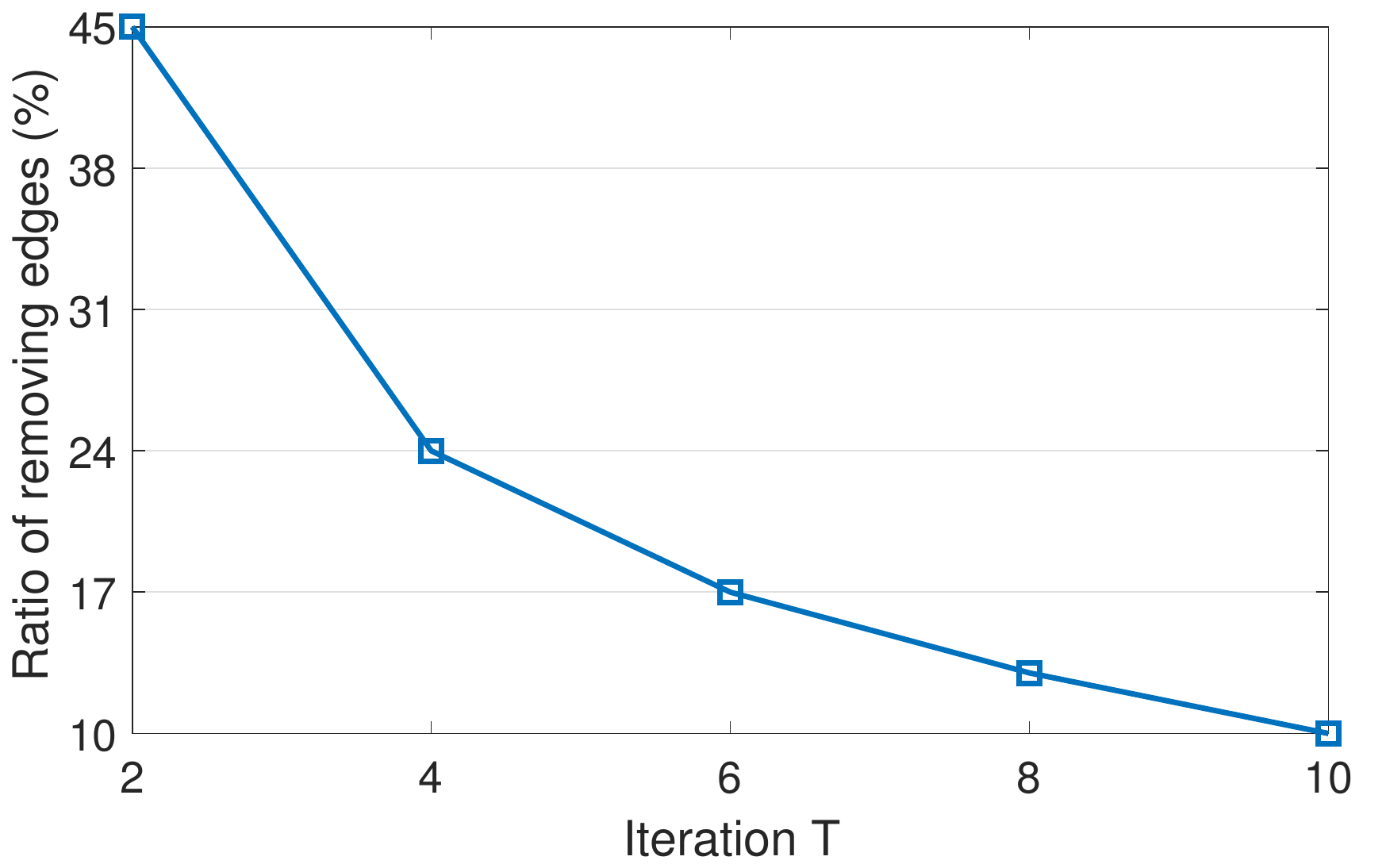}
		\end{minipage}
	}
	\caption{Parameter analysis of $\mu$ and $T$ on Cora and Citeseer datasets. }\label{fig::mu_T}
\end{figure*}
%
\subsection{Parameter Analysis} 
As shown in Eqs.(14,16), the sparsity of learned attention coefficients is controlled  
via parameters $\mu$ and $T$. 
In this section, we investigate the effect of parameter $\mu$ and iteration $T$ on affecting the sparsity of the learned attentions. 
To be specific, in our GSAT, parameters $\mu$ and $T$ control the ratio of removing edges and thus control the sparsity of attentions. Larger $\mu$ and smaller $T$ encourage to remove more edges and thus generate more sparse attentions. 
Figure \ref{fig::mu_T} shows the results across different values of $\mu$ and  $T$ on Cora and Citeseer datasets respectively. 
We can find that as parameter $\mu$ increases or $T$ decreases, the ratio of removing edges increases. This intuitively
demonstrates the desired sparsification behavior of our proposed spiking attention learning in our GSAT.


\section{Conclusion}
In this paper, we propose to leverage a Spiking Neural Network to learn graph attention for graph neural network. 
In contrast to  traditional self-attention mechanism used in existing GATs, the proposed spiking attention mechanism performs inexpensive computation by dealing with discrete spike trains and also returns sparse output. 
Based on the proposed spiking attention mechanism, we then develop a novel graph attention neural network model, termed Graph Spiking Attention network (GSAT).   Our GSAT can thus return sparse attention coefficients in natural and further conduct feature aggregation on the selective neighbors. Experimental results on several datasets demonstrate the effectiveness, energy-efficiency and robustness of our proposed GSAT model.


\begin{thebibliography}{10}
\bibitem{IF}
Larry~F Abbott.
\newblock Lapicque’s introduction of the integrate-and-fire model neuron
(1907).
\newblock {\em Brain research bulletin}, 50(5-6):303--304, 1999.

\bibitem{gatv2}
Shaked Brody, Uri Alon, and Eran Yahav.
\newblock How attentive are graph attention networks?
\newblock In {\em International Conference on Learning Representations}, 2022.

\bibitem{IFreview}
Anthony~N Burkitt.
\newblock A review of the integrate-and-fire neuron model: I. homogeneous
synaptic input.
\newblock {\em Biological cybernetics}, 95(1):1--19, 2006.

\bibitem{chaudhari2021attentive}
Sneha Chaudhari, Varun Mithal, Gungor Polatkan, and Rohan Ramanath.
\newblock An attentive survey of attention models.
\newblock {\em ACM Transactions on Intelligent Systems and Technology (TIST)},
12(5):1--32, 2021.

\bibitem{Cordone_2021_IJCNN}
Loic Cordone, Benoît Miramond, and Sonia Ferrante.
\newblock Learning from event cameras with sparse spiking convolutional neural
networks.
\newblock In {\em Proceedings of the IEEE International Joint Conference on
	Neural Networks (IJCNN)}, July 2021.

\bibitem{SpikingJelly}
Wei Fang, Yanqi Chen, Jianhao Ding, Ding Chen, Zhaofei Yu, Huihui Zhou,
Yonghong Tian, and other contributors.
\newblock Spikingjelly.
\newblock \url{https://github.com/fangwei123456/spikingjelly}, 2020.

\bibitem{fang2021deep}
Wei Fang, Zhaofei Yu, Yanqi Chen, Tiejun Huang, Timoth{\'e}e Masquelier, and
Yonghong Tian.
\newblock Deep residual learning in spiking neural networks.
\newblock {\em Advances in Neural Information Processing Systems},
34:21056--21069, 2021.

\bibitem{fang2021incorporating}
Wei Fang, Zhaofei Yu, Yanqi Chen, Timoth{\'e}e Masquelier, Tiejun Huang, and
Yonghong Tian.
\newblock Incorporating learnable membrane time constant to enhance learning of
spiking neural networks.
\newblock In {\em Proceedings of the IEEE/CVF International Conference on
	Computer Vision}, pages 2661--2671, 2021.

\bibitem{feng2022building}
Yifei Feng, Shijia Geng, Jianjun Chu, Zhaoji Fu, and Shenda Hong.
\newblock Building and training a deep spiking neural network for ecg
classification.
\newblock {\em Biomedical Signal Processing and Control}, 77:103749, 2022.

\bibitem{GeislerZG20}
Simon Geisler, Daniel Z\"{u}gner, and Stephan G\"{u}nnemann.
\newblock Reliable graph neural networks via robust aggregation.
\newblock In {\em Advances in Neural Information Processing Systems}, pages
13272--13284, 2020.

\bibitem{ghosh2009spiking}
Samanwoy Ghosh-Dastidar and Hojjat Adeli.
\newblock Spiking neural networks.
\newblock {\em International journal of neural systems}, 19(04):295--308, 2009.

\bibitem{glorot2010understanding}
Xavier Glorot and Yoshua Bengio.
\newblock Understanding the difficulty of training deep feedforward neural
networks.
\newblock In {\em Proceedings of the thirteenth international conference on
	artificial intelligence and statistics}, pages 249--256, 2010.

\bibitem{hamilton2017inductive}
Will Hamilton, Zhitao Ying, and Jure Leskovec.
\newblock Inductive representation learning on large graphs.
\newblock In {\em Advances in Neural Information Processing Systems}, pages
1024--1034, 2017.

\bibitem{han2020rmp}
Bing Han, Gopalakrishnan Srinivasan, and Kaushik Roy.
\newblock Rmp-snn: Residual membrane potential neuron for enabling deeper
high-accuracy and low-latency spiking neural network.
\newblock In {\em Proceedings of the IEEE/CVF conference on computer vision and
	pattern recognition}, pages 13558--13567, 2020.

\bibitem{cat}
Tiantian He, L~Bai, and Yew~Soon Ong.
\newblock Learning conjoint attentions for graph neural nets.
\newblock In {\em Advances in Neural Information Processing Systems (NeurIPS)
	34}. Curran Associates, Inc., 2021.

\bibitem{hunger2005floating}
Raphael Hunger.
\newblock {\em Floating point operations in matrix-vector calculus}, volume
2019.
\newblock Munich University of Technology, Inst. for Circuit Theory and
Signal~…, 2005.

\bibitem{GeCN}
Bo~Jiang, Beibei Wang, Jin Tang, and Bin Luo.
\newblock Gecns: Graph elastic convolutional networks for data representation.
\newblock {\em IEEE Transactions on Pattern Analysis and Machine Intelligence},
44(9):4935--4947, 2022.

\bibitem{prognn}
Wei Jin, Yao Ma, Xiaorui Liu, Xianfeng Tang, Suhang Wang, and Jiliang Tang.
\newblock Graph structure learning for robust graph neural networks.
\newblock In {\em Proceedings of the 26th ACM SIGKDD International Conference
	on Knowledge Discovery \& Data Mining}, pages 66--74, 2020.

\bibitem{kim2021how}
Dongkwan Kim and Alice Oh.
\newblock How to find your friendly neighborhood: Graph attention design with
self-supervision.
\newblock In {\em International Conference on Learning Representations}, 2021.

\bibitem{Adam}
Diederik~P Kingma and Jimmy Ba.
\newblock Adam: A method for stochastic optimization.
\newblock In {\em International Conference on Learning Representations}, 2015.

\bibitem{kipf2017semi}
Thomas~N. Kipf and Max Welling.
\newblock Semi-supervised classification with graph convolutional networks.
\newblock In {\em International Conference on Learning Representations (ICLR)},
2017.

\bibitem{ledinauskas2020training}
Eimantas Ledinauskas, Julius Ruseckas, Alfonsas Jur{\v{s}}{\.e}nas, and
Giedrius Bura{\v{c}}as.
\newblock Training deep spiking neural networks.
\newblock {\em arXiv preprint arXiv:2006.04436}, 2020.

\bibitem{attention_survey}
John~Boaz Lee, Ryan~A Rossi, Sungchul Kim, Nesreen~K Ahmed, and Eunyee Koh.
\newblock Attention models in graphs: A survey.
\newblock {\em ACM Transactions on Knowledge Discovery from Data (TKDD)},
13(6):1--25, 2019.

\bibitem{li2020deeprobust}
Yaxin Li, Wei Jin, Han Xu, and Jiliang Tang.
\newblock Deeprobust: A pytorch library for adversarial attacks and defenses.
\newblock {\em arXiv preprint arXiv:2005.06149}, 2020.

\bibitem{pmlr-v162-na22a}
Byunggook Na, Jisoo Mok, Seongsik Park, Dongjin Lee, Hyeokjun Choe, and Sungroh
Yoon.
\newblock {A}uto{SNN}: Towards energy-efficient spiking neural networks.
\newblock In {\em Proceedings of the 39th International Conference on Machine
	Learning}, pages 16253--16269, 2022.

\bibitem{pfeiffer2018deep}
Michael Pfeiffer and Thomas Pfeil.
\newblock Deep learning with spiking neurons: opportunities and challenges.
\newblock {\em Frontiers in neuroscience}, page 774, 2018.

\bibitem{ponulak2011introduction}
Filip Ponulak and Andrzej Kasinski.
\newblock Introduction to spiking neural networks: Information processing,
learning and applications.
\newblock {\em Acta neurobiologiae experimentalis}, 71(4):409--433, 2011.

\bibitem{9556508}
Nitin Rathi and Kaushik Roy.
\newblock Diet-snn: A low-latency spiking neural network with direct input
encoding and leakage and threshold optimization.
\newblock {\em IEEE Transactions on Neural Networks and Learning Systems},
pages 1--9, 2021.

\bibitem{dropedge}
Yu~Rong, Wenbing Huang, Tingyang Xu, and Junzhou Huang.
\newblock Dropedge: Towards deep graph convolutional networks on node
classification.
\newblock In {\em International Conference on Learning Representations}, 2020.

\bibitem{sen2008collective}
Prithviraj Sen, Galileo Namata, Mustafa Bilgic, Lise Getoor, Brian Galligher,
and Tina Eliassi-Rad.
\newblock Collective classification in network data.
\newblock {\em AI magazine}, 29(3):93--93, 2008.

\bibitem{sharmin2020inherent}
Saima Sharmin, Nitin Rathi, Priyadarshini Panda, and Kaushik Roy.
\newblock Inherent adversarial robustness of deep spiking neural networks:
Effects of discrete input encoding and non-linear activations.
\newblock In {\em European Conference on Computer Vision}, pages 399--414.
Springer, 2020.

\bibitem{shchur2018pitfalls}
Oleksandr Shchur, Maximilian Mumme, Aleksandar Bojchevski, and Stephan
G{\"u}nnemann.
\newblock Pitfalls of graph neural network evaluation.
\newblock {\em arXiv preprint arXiv:1811.05868}, 2018.

\bibitem{tavanaei2019deep}
Amirhossein Tavanaei, Masoud Ghodrati, Saeed~Reza Kheradpisheh, Timoth{\'e}e
Masquelier, and Anthony Maida.
\newblock Deep learning in spiking neural networks.
\newblock {\em Neural networks}, 111:47--63, 2019.

\bibitem{vaswani2017attention}
Ashish Vaswani, Noam Shazeer, Niki Parmar, Jakob Uszkoreit, Llion Jones,
Aidan~N Gomez, {\L}ukasz Kaiser, and Illia Polosukhin.
\newblock Attention is all you need.
\newblock {\em Advances in neural information processing systems}, 30, 2017.

\bibitem{velickovic2018graph}
Petar Veli{\v{c}}kovi{\'{c}}, Guillem Cucurull, Arantxa Casanova, Adriana
Romero, Pietro Li{\`{o}}, and Yoshua Bengio.
\newblock {Graph Attention Networks}.
\newblock {\em International Conference on Learning Representations}, 2018.

\bibitem{vreeken2003spiking}
Jilles Vreeken et~al.
\newblock Spiking neural networks, an introduction.
\newblock 2003.

\bibitem{ijcai2021-425}
Guangtao Wang, Rex Ying, Jing Huang, and Jure Leskovec.
\newblock Multi-hop attention graph neural networks.
\newblock In {\em Proceedings of the Thirtieth International Joint Conference
	on Artificial Intelligence, {IJCAI-21}}, pages 3089--3096. International
Joint Conferences on Artificial Intelligence Organization, 2021.

\bibitem{wang2019heterogeneous}
Xiao Wang, Houye Ji, Chuan Shi, Bai Wang, Yanfang Ye, Peng Cui, and Philip~S
Yu.
\newblock Heterogeneous graph attention network.
\newblock In {\em The world wide web conference}, pages 2022--2032, 2019.

\bibitem{SGAT}
Yang Ye and Shihao Ji.
\newblock Sparse graph attention networks.
\newblock {\em IEEE Transactions on Knowledge and Data Engineering}, pages
1--1, 2021.

\bibitem{neural_sparse}
Cheng Zheng, Bo~Zong, Wei Cheng, Dongjin Song, Jingchao Ni, Wenchao Yu, Haifeng
Chen, and Wei Wang.
\newblock Robust graph representation learning via neural sparsification.
\newblock In {\em International Conference on Machine Learning}, pages
11458--11468. PMLR, 2020.

\bibitem{spikgnn}
Zulun Zhu, Jiaying Peng, Jintang Li, Liang Chen, Qi~Yu, and Siqiang Luo.
\newblock Spiking graph convolutional networks.
\newblock In {\em Proceedings of the Thirty-First International Joint
	Conference on Artificial Intelligence, {IJCAI-22}}, pages 2434--2440, 2022.

\bibitem{zugner2018adversarial}
Daniel Z{\"u}gner, Amir Akbarnejad, and Stephan G{\"u}nnemann.
\newblock Adversarial attacks on neural networks for graph data.
\newblock In {\em ACM SIGKDD International Conference on Knowledge Discovery \&
	Data Mining}, pages 2847--2856, 2018.
\end{thebibliography}

\end{document}